\documentclass{article} 
\usepackage{colm2024_conference}

\usepackage{microtype}
\usepackage{hyperref}
\usepackage{url}
\usepackage{booktabs}
\usepackage{adjustbox}

\definecolor{darkblue}{rgb}{0, 0, 0.5}
\hypersetup{colorlinks=true, citecolor=darkblue, linkcolor=darkblue, urlcolor=darkblue}

\colmfinalcopy

\title{NeMo-Aligner: Scalable Toolkit for Efficient Model Alignment}

\author{Gerald Shen, Zhilin Wang, Olivier Delalleau, Jiaqi Zeng, Yi Dong,\\
\textbf{Daniel Egert, Shengyang Sun, Jimmy Zhang, Sahil Jain, Ali Taghibakhshi } \\
\textbf{Markel Sanz Ausin, Ashwath Aithal, Oleksii Kuchaiev} \\
NVIDIA \\
  \texttt{\{geshen, zhilinw\}@nvidia.com} 
  }

\begin{document}

\maketitle

\begin{abstract}
 
Aligning Large Language Models (LLMs) with human values and preferences is essential for making them helpful and safe.
However, building efficient tools to perform alignment can be challenging, especially for the largest and most competent LLMs which often contain tens or hundreds of billions of parameters. 
We create NeMo-Aligner, a toolkit for model alignment that can efficiently scale to a thousand GPUs for training the largest open-source LLMs such as Nemotron 4 340B and Llama 3.1 405B. 
NeMo-Aligner comes with highly optimized and scalable implementations for major paradigms of model alignment such as: Reinforcement Learning from Human Feedback (RLHF), Direct Preference Optimization (DPO), SteerLM,  and Self-Play Fine-Tuning (SPIN).
Additionally, our toolkit supports running most of the alignment techniques in a Parameter Efficient Fine-Tuning (PEFT) setting.
NeMo-Aligner is designed for extensibility, allowing support for other alignment techniques with minimal effort.
It is open-sourced with Apache 2.0 License and we invite community contributions at \url{https://github.com/NVIDIA/NeMo-Aligner}. 

\end{abstract}

\section{Introduction}

Pre-training large language models on tremendous amounts of unlabelled text has showcased promising capabilities \citep{brown2020language, zhang2022opt}. While such unsupervised pre-trained models have achieved impressive results, subsequently aligning models to follow user instructions is a  critical step to tap the capabilities of LLMs for practical use cases \citep{sanh2022multitask, wei2022finetuned}. Attempts based on Supervised Finetuning \citep{Conover_Hayes_2023,  kopf2023openassistant, alpaca} proved less effective compared to techniques that also made use of feedback to tune models towards responses that are more helpful and away from responses that are less so \citep{bai2022training, instructgpt, touvron2023llama, dong-etal-2023-steerlm}.

Despite the benefits of training models using feedback, these pipelines are notoriously challenging to get right \citep{lambert2023alignment, zheng2023secrets}, deterring widespread, productive adoption outside of select well-resourced organizations. For example, the popular Proximal Policy Optimization (PPO) variant of Reinforcement Learning from Human Feedback~(RLHF) approach \citep{instructgpt} requires running a complicated pipeline with four large language models interacting in a complex manner during training. Such alignment algorithms introduce new system challenges for efficient training that require re-thinking various aspects of the software stack including model scalability, coordination among models, and text generation within the training loop.

There are existing open source tools for model alignment, most notably HuggingFace TRL~\citep{vonwerra2022trl}, CarperAI trlX~\citep{havrilla-etal-2023-trlx} and Microsoft DeepSpeed-Chat~\citep{yao2023dschat}.
These tools provide an excellent starting point with respect to usability and feature set. However, with NeMo-Aligner we aim to vastly improve performance and scalability to more than a thousand GPUs, especially useful for aligning the largest and most competent models such as Nemotron 4 340B \citep{nvidia2024nemotron4340btechnicalreport}, Llama 3.1 405B \citep{llama3.1} and beyond.

NeMo-Aligner addresses scalability challenges by (I)~building upon Megatron-LM~\citep{shoeybi2020megatronlm} with 3D (data, tensor, and pipeline)-parallelism training, (II)~having a distributed approach to Proximal Policy Optimization (PPO) training in RLHF and (III)~integrating PPO inference optimizations based on TensorRT-LLM \citep{trtllm} during rollout stage. Combined, these optimizations allow users to efficiently train the largest models over a thousand GPUs reducing research iteration time. 


NeMo-Aligner optimizes popular alignment techniques including Supervised Finetuning (SFT), PPO-based RLHF \citep{instructgpt}, Direct Preference Optimization \citep{rafailov2023direct}, SteerLM~\citep{dong-etal-2023-steerlm} and Self-Play Fine-Tuning \citep{chen2024selfplay}. We briefly outline the background for these techniques in Section \ref{sec:background}, followed by an in-depth exploration of training with each of the techniques in Sections \ref{sec:rlhf_training}, 
\ref{sec:dpo_training},
\ref{sec:steerlm_training}, 
and \ref{sec:spin_training}. Finally,
we demonstrate the extensible design of NeMo-Aligner in Section \ref{sec:framework_extensibility}.

\section{Model Alignment Background}\label{sec:background}

\begin{figure*}[ht]
\centering
\includegraphics[width=\textwidth,]{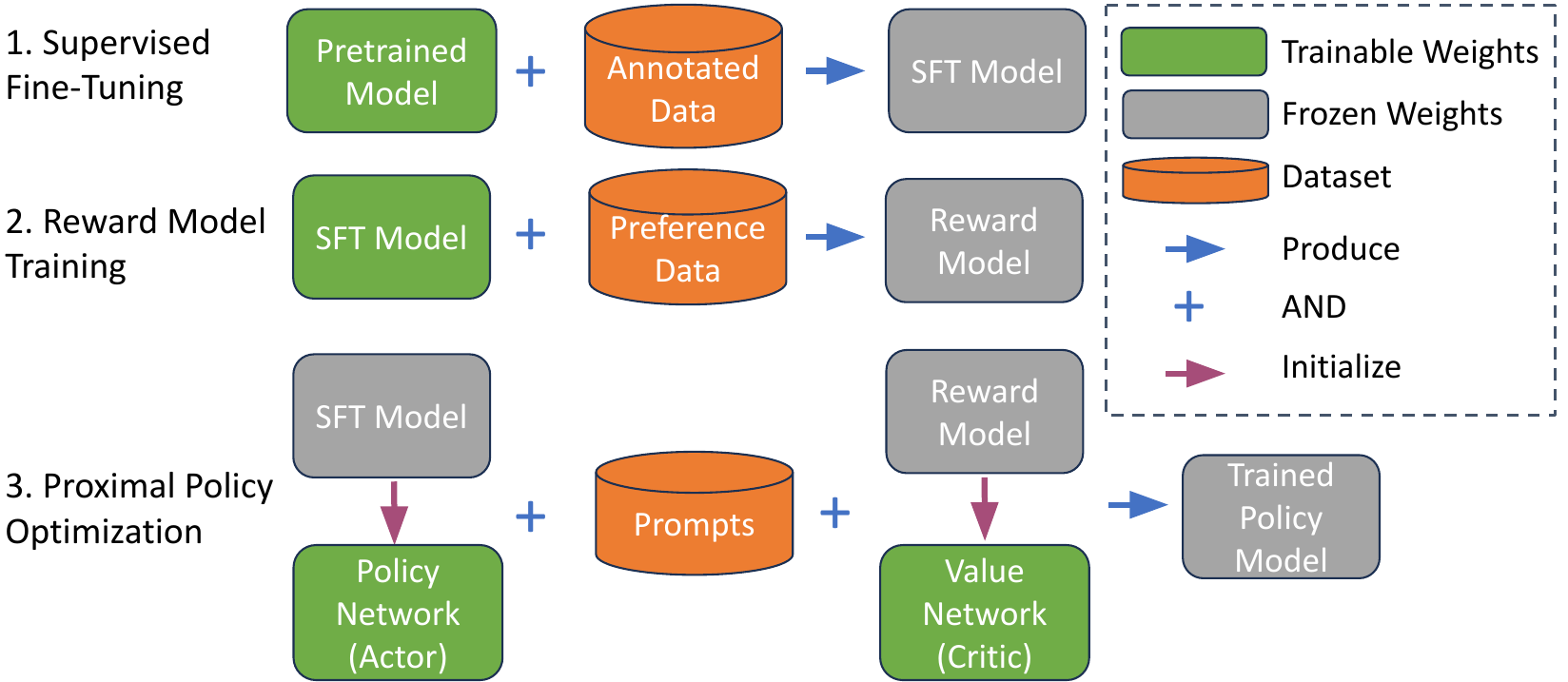}
\caption{Training Recipe for RLHF based on \cite{instructgpt}. \textit{Step 1}:~Annotated Prompt-Response Data is used to perform Supervised Fine-Tuning on the pre-trained (base) Model. \textit{Step 2}:~The resulting SFT model is trained with Preference Data to produce a Reward Model. \textit{Step~3}:~The SFT Model is used to initialize the Policy Network, and the Reward Model is used to initialize the Value Network -- together with input prompts, all four models are used to train a Policy Model. The SFT model is also used to compute the KL divergence penalty in Step 3 (not illustrated).}
\label{fig:rlhf-workflow}
\end{figure*}

\subsection{Supervised Fine Tuning}\label{sec:sft}

Given a pre-trained (also referred to as "base") model, supervised fine-tuning~(SFT) updates the base model's parameters on prompts with expected responses, where the expected responses might come from expert human annotations \citep{kopf2023openassistant} or other language models \citep{ding2023enhancing}. The model is trained to mimic the expected responses given prompts using the token-level cross-entropy loss. SFT is an important prerequisite step in Reinforcement Learning from Human Feedback \citep{instructgpt} and Direct Preference Optimization \citep{rafailov2023direct} because without it, the base model is very unlikely to generate responses which follow user's instructions. This step is also sometimes called \textit{behavior cloning} because the model is expected to mimic responses of a human or another model.

\subsection{Reinforcement Learning from Human Feedback}
\label{sec:rlhf}

Reinforcement Learning from Human Feedback~(RLHF) was introduced by~\citet{christiano2017rlhf} as a way to avoid manually defined reward functions in Reinforcement Learning.
Instead, a reward model is trained from a dataset of human preferences consisting of pairs of ``chosen'' and ``rejected'' trajectories.
The reward model's loss, derived from the Bradley-Terry model~\citep{bradleyterry1952}, tries to maximize the likelihood that $r_{chosen} > r_{rejected}$ (i.e., that the predicted rewards are consistent with human preferences).
Once the reward model is trained, it may be used to compute rewards for RL algorithm. Two most common methods used in RLHF are REINFORCE \citep{williams1992simple} and Proximal Policy Optimization (PPO) \citep{schulman2017proximal}. In NeMo-Aligner we focus on PPO, specifically as described by \citet{instructgpt}.

RLHF has been shown to bring significant benefits for model alignment~\citep{instructgpt,bai2022training,touvron2023llama} with the typical training recipe being as follows, also illustrated in Figure \ref{fig:rlhf-workflow}:
\begin{enumerate}
    \item From a pre-trained base model, train an initial SFT model as described in Section \ref{sec:sft}.
    \item \label{step:sft} From the SFT model, train a reward model using a dataset of human preferences made of pairs of ``chosen'' and ``rejected'' responses to a set of prompts, following~\citet{christiano2017rlhf}. Typically, we initialize a linear reward model head on top of the SFT model before training.
    \item \label{step:rlhf} From the SFT model, train a policy with the online Proximal Policy Optimization algorithm~(PPO, \citealp{schulman2017proximal}), with rewards provided by the trained reward model. Input prompts may not necessarily be the same as those used for reward model training. A regularization term based on the KL divergence w.r.t. the SFT model helps prevent the policy from straying too far away from its starting point and exploiting the ``blind spots'' of the reward model~\citep{stiennon2020summarize,instructgpt}.
    The PPO critic is typically initialized from the reward model.
\end{enumerate}

\subsection{Direct Preference Optimization}\label{sec:dpo}
Direct Preference Optimization \citep{rafailov2023direct} is an offline, off-policy algorithm that makes use of preference data to directly train an optimal policy without an explicit reward model. Rather than use a reward model, a reference policy is used to implicitly derive the reward between a chosen and rejected pair via the Bradley-Terry model. This is accomplished via the difference in the log probabilities between the chosen and rejected responses, which is calculated for the optimal and reference policies. This difference is scaled and then transformed by the sigmoid function to derive the loss. The reference policy is frozen during training and represents the policy used to generate the chosen/rejected responses to the given prompts. If the reference policy used to generate the preference data is not available, it can be approximated by supervised fine-tuning on the prompts and preferred responses of the preference data.

\subsection{SteerLM}

SteerLM \citep{dong-etal-2023-steerlm} is a model alignment algorithm based on supervised finetuning which avoids use of complex RL methods, similarly to DPO. SteerLM involves three steps. The first step is to train an Attribute Prediction Model that learns to predict the values~(between 0 and 4 where higher is more) for various semantic aspects of a response that make responses helpful and safe, such as its correctness and toxicity \citep{kopf2023openassistant, wang2023helpsteer}. Next, the Attribution Prediction Model can be used to annotate the various attributes contributing to helpfulness and safety in a diversity of prompt-response datasets. Finally, these annotated datasets can be used to perform Attribute-Conditioned Supervised Fine-Tuning where the model learns to generate the response conditioned on the prompt as well as the annotated attributes formatted into a string, such as \texttt{helpfulness:4,correctness:4,toxicity:0}. This step teaches the model to discriminate between responses that are more helpful/safe and those that are less, in a fine-grained manner for each semantic aspect. At inference time, the prompt can be appended with the optimal attribute values, as above, to generate the most helpful response.

\subsection{Self-Play Fine-Tuning}
Self-Play Fine-Tuning (SPIN) \citep{chen2024selfplay} is a self-play based algorithm, where a strong model is developed from a weaker model by playing against previous instances of itself. Starting from an SFT dataset of prompt/response pairs, new responses are generated from previous iterations of the model. Its policy is then improved by discriminating between these self-generated responses and the ground truth human-generated SFT responses. This is accomplished through a preference loss function which is identical to the one used by DPO (Section \ref{sec:dpo}). When SPIN training first starts, we use a copy of the initial policy as the reference policy in the DPO loss. The self-play ``game'' is then played for a number of iterations during which we train the policy as in DPO whilst keeping the reference policy frozen, and at the end of each iteration we update the reference policy's weights with those from the trained policy. During each iteration, we iterate over our SFT training dataset and use the reference policy to generate responses for each prompt, building a preference tuple between the ground truth SFT human ``chosen'' response and the generated ``rejected'' response. Once we have these preference tuples for the entire epoch, we update the model weights via the DPO loss function from these tuples of ``(chosen, rejected)'' preference pairs. The model thus implicitly learns to prefer the ground truth SFT responses to those generated by the previous iteration of itself, which forms the self-play mechanism.

\section{RLHF (PPO) Training}\label{sec:rlhf_training}

\begin{figure}[h]
\centering
\includegraphics[width=1\textwidth,]{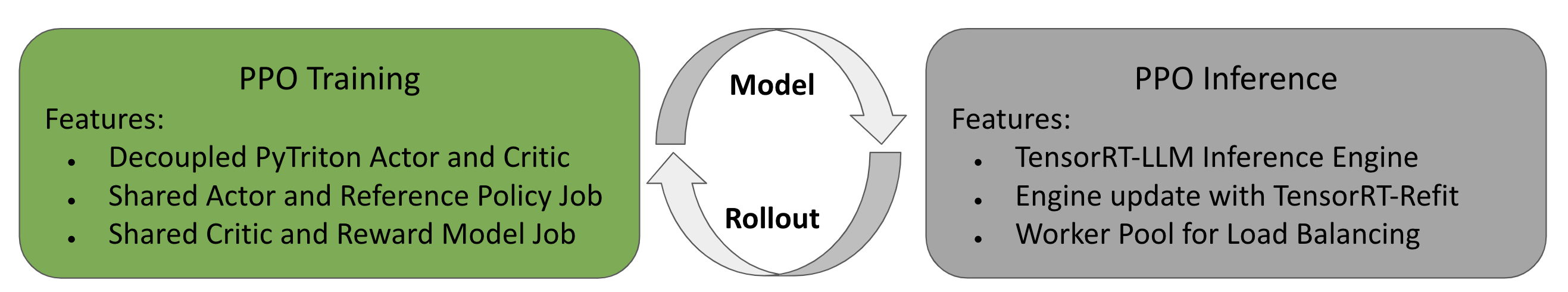}
\caption{Optimizations for RLHF training. Optimizations for PPO training and inference are detailed in Sections \ref{sec:ppo_training} and \ref{sec:ppo_inference} respectively.
}
\label{fig:example3}
\end{figure}

NeMo-Aligner is designed to support numerous alignment techniques efficiently at extremely large scales. It does so by building upon Megatron-LM \citep{shoeybi2020megatronlm} and NeMo \citep{nemo} to include features such as optimized kernels from Transformer Engine \citep{te}, distributed fused adam optimizer and 3D parallelism support. NeMo-Aligner supports the entire RLHF pipeline as introduced by \citet{instructgpt} and described in Section~\ref{sec:rlhf}. The training pipeline is separated into three distinct stages as illustrated in Figure \ref{fig:rlhf-workflow}: Supervised Fine-Tuning, Reward Model Training, and Proximal Policy Optimization. The challenges with the pipeline efficiency come primarily from the Proximal Policy Optimization stage, and this section describes our approach to tackling these challenges, as summarized in Figure~\ref{fig:example3}.

\subsection{Distributed Approach to PPO training}\label{sec:ppo_training}
\begin{figure}[h]
\centering
\includegraphics[width=1\textwidth,]{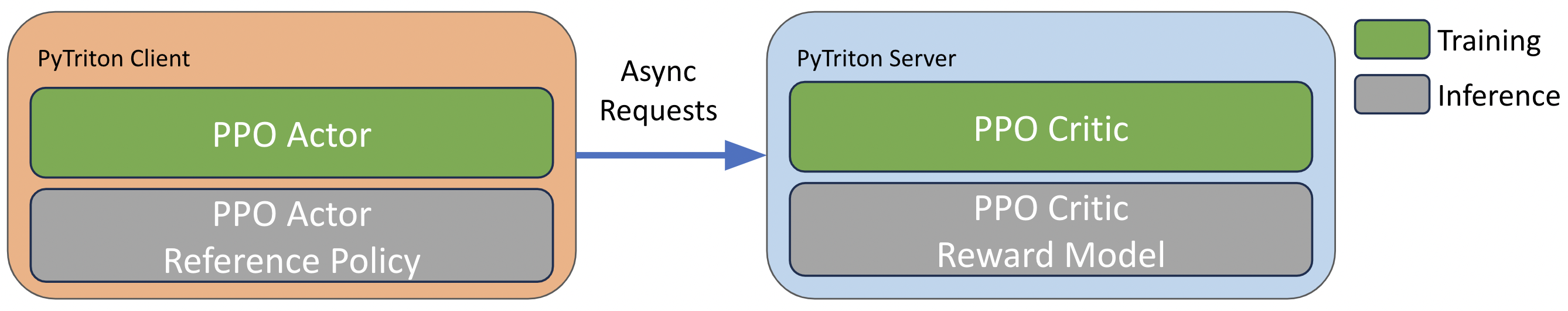}
\caption{NeMo-Aligner PPO System Architecture. The PPO Actor is a PyTriton~\citep{pytriton} client that sends async requests to the server~(PPO critic and reward model) to obtain the rewards and values of generated rollouts, and to send the training data for the critic.}
\label{fig:example2}
\end{figure}

The PPO stage requires running training and/or inference on four different models, as illustrated in Figure \ref{fig:example2}:
\begin{enumerate}
    \item PPO Actor~(training and inference, initialized from SFT model): The model we want to fine tune with PPO.
    \item Reference Policy~(inference only, set to the SFT model): The model to compute the KL penalty against.
    \item PPO Critic~(training and inference, initialized from the reward model): Used in PPO to compute value estimates.
    \item Reward Model~(inference only) : Provides RL rewards on generated rollout data.
\end{enumerate}

All of these models can be extremely large (e.g. Llama 3.1 405B), so NeMo-Aligner takes a distributed approach to PPO training. We allow users to setup PyTriton \citep{pytriton} servers and clients to communicate across the different models during PPO. These PyTriton servers make it possible to run the models on different compute clusters, removing the requirement of having both the critic and actor on the same compute allocation. Naively, four different servers (i.e. one for each model) would be launched. However, we note that the reference policy and PPO actor are the same model but with different weights. Therefore, we combine them into one job and offload the reference policy's weights to CPU, swapping them with the actor's weights for the reference policy inference step. We deploy the same strategy for the reward model and critic. All communications are done asynchronously, permitting pipelined critic inference/training with policy inference/training. 

We scale compute allocation sizes such that the [reward model inference + critic inference] $\approx$ [actor sampling + reference policy inference] and [critic train] $\leq$ [actor train + actor inference initialization]. This ensures that the pipeline can use available compute capacity most efficiently.

\subsection{Optimizations for PPO rollout}\label{sec:ppo_inference}

Response generation during the rollout step dominates the end to end time of PPO training. The generation stage of the actor is composed of multiple forward passes, with one token generated per forward pass. Therefore, generation stage kernels are generally launch latency and memory bandwidth bound, meaning that directly reusing the compute optimized forward pass implementation of the training stage results in very poor performance.

To address these bottlenecks, we implement the generation stage using TensorRT-LLM \citep{trtllm}, a high-performance LLM deployment framework. TensorRT-LLM integrates inference optimized kernels and automatic kernel fusion into a TensorRT based runtime to achieve better performance.
At the start of RLHF, the model is passed to TensorRT-LLM which compiles the model into a TensorRT engine; TensorRT-LLM loads the engine into its runtime and performs generation. The engine holds a copy of the model weights along with the runtime KV-cache and activations. Due to the cost of serializing the engine, we keep the engine in memory during training. As a result, we reduce the peak memory pressure by recomputing training stage activations in the backward pass. Furthermore, because generation has lower memory requirements than training, we reshard the model to only use tensor parallelism during inference if memory allows, removing overhead from inter-node communication when otherwise running with pipeline parallelism.

On subsequent training steps, the engine must be synced with updated parameter weights from the training stage. The engine is updated in-place using the TensorRT Refitter \citep{trtrefitter}. We avoid recompiling the engine which would incur a large overhead as generation cannot begin until the weights are updated.

For large problem sizes, there may be discrepancies of generation time between the fastest and the slowest data parallel worker during generation due to the differences in response lengths.
To mitigate this, we allow users to setup a worker pool to dynamically load balance among data parallel workers to give workers with shorter generations correspondingly more work.

\subsection{Model Training Details and Quality}\label{sec:rlhf_details}
 As a demonstration of practical large-scale RLHF training with NeMo-Aligner, we train Llama3 \citep{llama3modelcard} 70B model using PPO as prescribed in \citet{helpsteer2}. The PPO model is trained with a rollout global batch size of 128, training global batch size of 128, constant learning rate of 1e-7, KL penalty of $0.003$, and using HelpSteer2 as a prompt source.

Following \citet{helpsteer2, meng2024simposimplepreferenceoptimization}, we use MT-Bench~\citep{zheng2023judging} with GPT-4-Turbo judge to evaluate the performance of the trained RLHF Model. The resulting model achieves a performance of 8.13 on MT-Bench, which is an improvement upon the strong starting SFT checkpoint with MT-Bench of 7.96. The Reward Model and RLHF-trained model are publicly released at \url{https://huggingface.co/nvidia/Llama3-70B-SteerLM-RM} and \url{https://huggingface.co/nvidia/Llama3-70B-PPO-Chat} respectively.

\subsection{Scalability}

\begin{table}[ht!]
\centering
\begin{tabular}{lccc}
\toprule
No. of compute nodes (Actor + Critic) & 8 + 4  & 16 + 8 \\
\midrule
\textit{Time per step in seconds (std.) $\downarrow$} \\
\midrule

Overall & 53.7 (2.223) & 29.8 (1.459) \\
\quad Train & 8.5 (0.356) & 4.8 (0.118) \\
\quad Rollout &  45.2 (1.957) & 25.0 (1.41) \\
\quad \quad - Response generation & 35.8 (1.658) & 17.5 (1.224) \\

\quad \quad - Log-probs calculation & 4.0 (0.573) & 2.8 (0.275) \\
\quad \quad - TensorRT Refit & 3.1 (0.182) & 2.3 (0.334) \\
\quad \quad - Critic wait & 0.01 (0.001) & 0.01 (0.001) \\

\midrule
\textit{Relative speed up (vs. 8 + 4 node setup) $\uparrow$} \\
\midrule
Overall & 1x & 1.80x \\ 
\quad Train & 1x & 1.77x \\ 
\quad Rollout & 1x & 1.81x \\
\quad \quad - Response generation & 1x & 2.05x \\
\quad \quad - Log-probs calculation & 1x & 1.43x \\

\bottomrule
\end{tabular}
\caption{Effects of scaling training across different number of  compute nodes for Llama 3 70B actor and Llama 3 70B critic on rollout global batch size of 128 and BF16 precision following Section \ref{sec:rlhf_details}. Nodes are 8*H100-80GB-SXM connected with intra-node NVLink \citep{nvlink} and inter-node Infiniband \citep{infiniband} interconnects. Time per step calculated based on mean of 5 steps after the first step, as the first step incurs substantial time for TRT-LLM Engine Building. Further training configuration details are in Table \ref{tab:parallelism}.
}
\label{tab:scaling}
\end{table}

To demonstrate the scaling efficiency of NeMo-Aligner, we repeat identical training setups from Section \ref{sec:rlhf_details}, with 8 actor nodes + 4 critic nodes and 16 actor nodes + 8 critic nodes. As shown in Table \ref{tab:scaling}, overall time per step reduces correspondingly, achieving a 1.80x speed up between 8+4 nodes and 16+8 nodes. The speed up in overall time per step is contributed by speed ups in both the Train and Rollout stages, demonstrating the effective optimization that NeMo-Aligner has done for both stages. 

The scaling of Train stage is sublinear due to number of micro-batches per data parallel rank decreasing as node count increases. Because all pipeline stages must complete before the optimizer is called in pipeline parallel models, we incur an overhead to fill and drain the pipeline that is independent of the number of micro-batches  \citep{shoeybi2020megatronlm}. Therefore, decreasing the number of micro-batches per data parallel rank increases the proportion of the train step spent in filling and draining the pipeline, where GPU utilization is poor. A similar issue is apparent during the log prob calculation phase with scaling of 1.43x.

\begin{table}[ht!]
\centering
\begin{tabular}{lcccc}
\toprule
No. of compute nodes (Actor + Critic) & 16 + 8 & 32 + 16 & 64 + 32 \\
\midrule
\textit{Time per step in seconds (std.) $\downarrow$} \\
\midrule

Overall & 190.4 (15.392) & 106.8 (6.842) & 56.9 (7.596) \\	
\quad Train & 38.8 (4.674) & 22.2 (2.408) & 14.1 (1.439) \\
\quad Rollout & 151.5 (13.172) & 84.6 (5.596) & 42.7 (6.36) \\
\quad \quad - Response generation & 131.3 (12.551) & 71.7 (5.093) & 29.9 (6.456) \\

\quad \quad - Log-probs calculation & 15.5 (3.055) & 8.5 (1.753) & 6.2 (1.177) \\
\quad \quad - TensorRT Refit & 2.2 (0.408) & 2.2 (0.014) & 3.5 (0.063) \\
\quad \quad - Critic wait & 0.02 (0) & 0.03 (0.001) & 1.7 (2.045) \\

\midrule
\textit{Relative speed up (vs. 16 + 8 node setup)$\uparrow$} \\
\midrule
Overall & 1x & 1.78x & 3.35x \\ 
\quad Train & 1x & 1.75x &  2.75x \\ 
\quad Rollout & 1x & 1.79x & 3.55x \\
\quad \quad - Response generation & 1x & 1.83x & 4.39x \\
\quad \quad - Log-probs calculation & 1x & 1.82x & 2.50x \\

\bottomrule
\end{tabular}
\caption{Effects of scaling training for Llama 3 70B actor and Llama 3 70B critic on rollout global batch size of 1024 and BF16 precision. Nodes are 8*H100-80GB-SXM connected with intra-node NVLink \citep{nvlink} and inter-node Infiniband \citep{infiniband} interconnects. Time per step calculated based on mean of 5 steps after the first step, as the first step incurs substantial additional time for TRT-LLM Engine Building. Our distributed optimizer feature enables NeMo-Aligner to support double the rollout microbatch size from 8 to 16 in the 64 + 32 node configuration. This speeds up the response generation time significantly. Further configuration details are in Table \ref{tab:parallelism}. 
}
\label{tab:scaling2}
\end{table}

Generation time scales well with the number of nodes, achieving a near-linear speedup of 2.05x. This is because scaling up the number of actor nodes proportionally increases the number of data parallel workers for each step, which can evenly share the intense work of generation.
However, the time spent fitting weights into the TensorRT engine remains relatively constant and therefore does not scale well with node count.
Finally, async communications between the Actor and the Critic models result in the additional time taken to wait for the Critic model to be inconsequential (0.01 seconds), suggesting the effectiveness of having async non-blocking calls between actor and critic models in the PPO pipeline. 

\begin{table}[ht!]
\begin{adjustbox}{max width=\columnwidth, scale=1}
\centering
\begin{tabular}{lccccccc}
\toprule
Model & No. of compute nodes & Tensor Parallel & Pipeline Parallel & Data Parallel & \multicolumn{2}{c}{Rollout Batch Size}\\
&&&&& Micro & Global \\

\midrule
70B + 70B & 8 + 4  & 8 + 8 & 8 + 4 & 1 + 1 & 8 & 128\\
70B + 70B & 16 + 8 & 8 + 8 & 8 + 4 & 2 + 2 & 8 & 128 / 1024\\
70B + 70B & 32 + 16 & 8 + 8 & 8 + 4 & 4 + 4 & 8 & 1024\\
70B + 70B & 64 + 32 & 8 + 8 & 8 + 4 & 8 + 8 & 16 & 1024\\
405B + 405B & 84 + 42 & 8 + 8 & 21 + 21 & 4 + 2 & 16 & 128\\
\bottomrule
\end{tabular}
\end{adjustbox}
\caption{Parallelism settings for scaling experiments on Llama3 models. The node counts and parallelism configurations are denoted as actor + critic.}
\label{tab:parallelism}
\end{table}

System scalability also needs to be considered under the context of the problem requirements. The training setup in Section \ref{sec:rlhf_details} has a 70B Llama 3 Actor, 70B Llama 3 Critic as well as a rollout global batch size of 128. Such a setup limits the effective demonstration of our system scaling beyond 16 + 8 nodes as there is not enough work to be meaningfully shared across more data parallel workers. Therefore, we modify the setup slightly to use a rollout global batch size of 1024 in Table \ref{tab:scaling2} in order to measure the system performance when the requirements are higher. Table \ref{tab:scaling2} shows that the increased requirements of the training job allows it to meaningfully scale to 64 + 32 nodes (with 768 H100 GPUs total) for various stages within PPO.

\subsection{What contributes to system performance?}

\begin{table}[ht!]
\centering
\begin{tabular}{lcc}
\toprule
 & Time per step  & Time relative to  \\
 & in seconds (std.) $\downarrow$ & Optimal RLHF setup $\downarrow$ \\

\midrule

Optimal RLHF Setup & 53.7 (2.223) & 1x \\
- TensorRT-LLM Integration & 372.2 (37.433) & 6.93x \\
  (\textit{i.e.} using NeMo Generate)\\
- Reshard & 207.6 (8.940) & 3.87x \\
- TensorRT Refit & 169.0 (5.206) & 3.15x \\
- Async Requests & 82.8 (7.826)  & 1.54x \\


\bottomrule
\end{tabular}
\caption{Ablation studies on training Llama 3 70B actor and critic on rollout global batch size of 128 with 8 Actor nodes and 4 critic nodes. Nodes are 8*H100-80GB-SXM connected with intra-node NVLink \citep{nvlink} and inter-node Infiniband \citep{infiniband} interconnects. Time per step is calculated based on mean of 5 steps after the first step, as the first step incurs substantial additional time for TRT-LLM Engine Building. }
\label{tab:ablation}
\end{table}

To better understand the importance of each aspect of NeMo-Aligner's PPO system design, we conduct ablation studies by removing one aspect at a time and measuring the overall time per step as shown in Table \ref{tab:ablation}. We find that TensorRT-LLM Integration is the most critical component for high system performance, without which PPO will take nearly seven times as long for each step. This is followed by resharding our model to use tensor parallelism only during inference (3.87x), using TensorRT Refit to avoid TensorRT-LLM engine recompiling (3.15x), the use of async requests between actor and critic models (1.54x). We did not observe meaningful speedup in using the worker pool to balance the work between data parallelism ranks because the problem size is small (with rollout global batch size of 128) and therefore the worker imbalance is less than the overhead we incur doing the balancing itself. Nevertheless, we expect this feature to help for larger problem sizes.

\subsection{Training the largest open source LLM}

\begin{table}[ht!]
\centering
\begin{tabular}{lc}
\toprule
No. of compute nodes (Actor + Critic) & 84 + 42 \\
\midrule
\textit{Time per step in seconds (std.) $\downarrow$} \\
\midrule
Overall & 164.60 (3.434) \\
\quad Train & 5.60 (0.532) \\
\quad Rollout & 158.90 (2.969) \\
\quad \quad - Response generation & 140.10 (1.067) \\
	
\quad \quad - Log-probs calculation & 17.20 (2.352) \\
\quad \quad - TensorRT Refit & 0.70 (0.108) \\
\quad \quad - Critic wait & 0.30 (0.454) \\

\bottomrule
\end{tabular}
\caption{NeMo Aligner supports the largest open source LLM as of July 2024 (Llama 3.1 405B) as both the actor and critic. We run the model on rollout global batch size of 128 and BF16 precision. Nodes are 8*H100-80GB-SXM connected with intra-node NVLink \citep{nvlink} and inter-node Infiniband \citep{infiniband} interconnects. Time per step is calculated based on mean of 5 steps after the first step, as the first step incurs substantial additional time for TRT-LLM Engine Building. Further configuration details are in Table \ref{tab:parallelism}.}
\label{tab:405b}
\end{table}

NeMo-Aligner supports alignment of the largest open source LLMs as of July 2024, such as Nemotron 4 340B \citep{nvidia2024nemotron4340btechnicalreport} and Llama 3.1 405B \citep{llama3.1}.
In Table \ref{tab:405b}, we perform PPO on Llama 3.1 405B using 1008 H100 GPUs, using a configuration based on Sec. \ref{sec:rlhf_details}. We use Llama 3.1 405B Instruct as the actor and we train a Reward Model on top of Llama 3.1 405B Instruct using the same data and hyper-parameters as Nemotron 4 340B Reward \citep{helpsteer2}. Compared to the 70B model, the 405B model is substantially slower to train during PPO, mainly bottle-necked by the slow response generation stage. This is because the 405B model generation cannot fit into a single node and requires pipeline-parallel generation which we plan to optimize further in future work. Additionally, in the first few steps of PPO, the 405B model generates longer responses (mean of 916 tokens) than the 70B model with a mean of 351 tokens.

\section{DPO Training}\label{sec:dpo_training}

We follow the Zephyr-7B-Beta \citep{tunstall2023zephyr} training recipe, a model trained with SFT and DPO. Briefly, SFT was first performed on Mistral-7B  \citep{jiang2023mistral} using the Ultrachat dataset \citep{ding2023enhancing}. Model was then further trained with DPO using the Ultrafeedback dataset \citep{cui2023ultrafeedback}. For SFT, we used a constant learning rate of $2e-5$, global batch size of $512$, and trained the model for $3$ epochs. For DPO training, we used KL regularization coefficient of $3e-4$, global batch size of $512$ and a cosine learning rate schedule with peak LR of $1e-7$, minimum LR of $1e-8$, $50$ warmup steps, and max. $300$ steps. We obtain slighter better MT-Bench scores than those reported by \citet{tunstall2023zephyr} for both the final model (7.60 vs 7.34) and the SFT-only initial model (6.77 vs 6.64). 

\section{SteerLM Training with LoRA}\label{sec:steerlm_training}

Low Rank Adaptation ~\citep{hu2021lora} enables fine-tuning large language models in a more efficient and cost-effective manner. Supported for various alignment techniques within NeMo-Aligner, LoRA is applied to SteerLM training following the training recipe by \citet{wang2023helpsteer} using the Llama~2 70B model as well as 
the HelpSteer \citep{wang2023helpsteer} and Open Assistant datasets \citep{kopf2023openassistant}. Specifically, we applied LoRA to all attention layers, with a rank of 32. We used global batch size of 128, constant learning rate of $1e-5$ after 10 warmup steps with the AdamW optimizer, and trained for 3 epochs. As shown in Table~\ref{tab:lora}, applying LoRA to SteerLM training with BF16 can reduce the minimum number of 80GB GPUs required from 32 to 8. With the same number of GPUs, LoRA achieves a $5\times$ speedup compared to full-parameter fine-tuning, while maintaining comparable model performance: MT-Bench 7.43 vs. 7.54, which is within noise level for this benchmark \citep{jiang2023mistral}.

\begin{table}[!h]
\centering
\begin{adjustbox}{max width=\columnwidth, scale=1}
\begin{tabular}{lcc}
\toprule
& \textit{Full-Param} & \textit{LoRA} \\
\midrule
$\#$ trainable params & 70B & 89M \\
$\min \#$ 80GB GPUs required& 32 & 8\\
Relative speed (sample/GPU/s) & $1\times$ & $5\times$ \\
MT-Bench & 7.54 & 7.43 \\
\bottomrule
\end{tabular}
\end{adjustbox}
\caption{Comparison of Full-Parameter and LoRA SteerLM following training recipe by \citet{wang2023helpsteer}.}
\label{tab:lora}
\end{table}

\begin{figure}[h]
\centering
\includegraphics[width=10cm,]{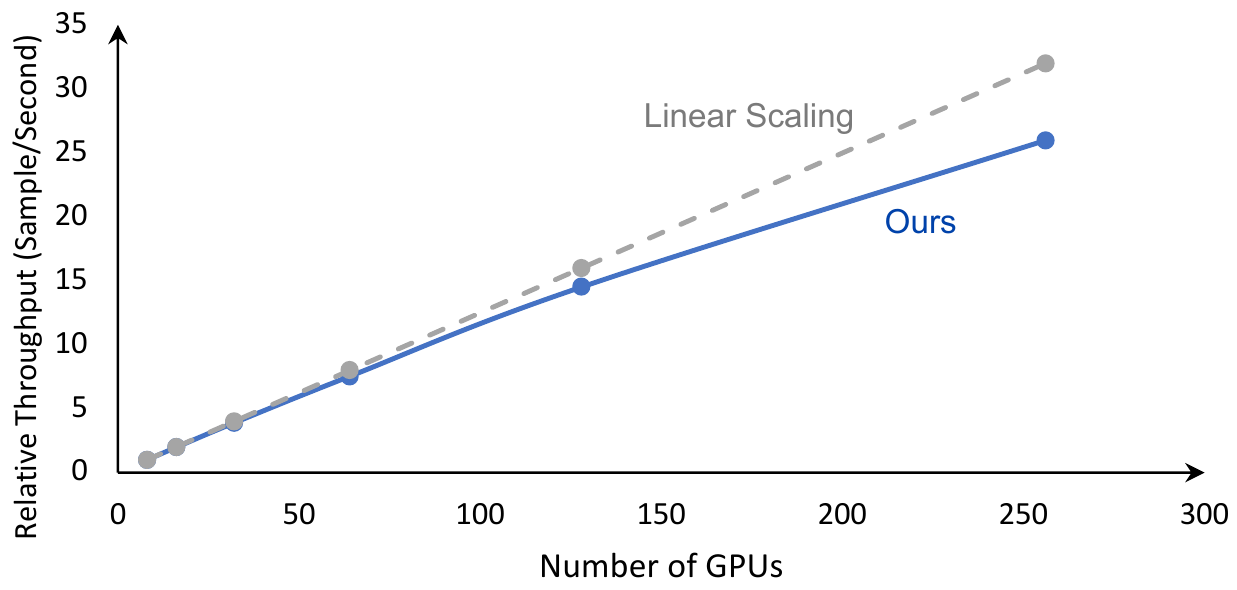}
\caption{Relative throughput of LoRA applied to SteerLM training as the number of GPUs increases.}
\label{fig:lora_scaling}
\end{figure}

As we increase the number of GPUs used for LoRA training, the relative throughput (measured in samples per second) improves almost proportionally, as shown in Figure~\ref{fig:lora_scaling}. This shows that NeMo-Aligner can effectively distribute and parallelize the workload across a large number of GPUs with minimal overhead and diminishing returns.

\section{SPIN Training}\label{sec:spin_training}

We recreate the Zephyr-7B-Beta \citep{tunstall2023zephyr} SFT model via SPIN instead of SFT as formulated by \citet{chen2024selfplay}. We start with the Mistral-7B base model \citep{jiang2023mistral} and perform SPIN training following \citet{chen2024selfplay}. However, we make a few departures from their methodology, in that we do not inject generations from the previous iteration into the current iteration (which would double the dataset size every epoch), and we only train for a single iteration, with 1 epoch per iteration. Additionally, we use a random subset of only 50k samples from Ultrachat200k \citep{ding2023enhancing} rather than the entire dataset, and use AdamW instead of RMSProp. Our learning rate is $5e-7$ with 400 total steps, 40 warmup steps, and this LR is then decayed to $1e-7$ for the last 100 steps using cosine annealing. Global batch size is 64, weight decay is 0.0, and the KL regularization coefficient is $0.1$, as per \citet{chen2024selfplay}. Using this approach, we achieve an MT-Bench score of 7.04 which exceeds the 6.64 of Zephyr-7B-Beta using SFT \citep{tunstall2023zephyr}, as well as the 6.78 of the 3-iteration SPIN model \citep{chen2024selfplay}. 

\section{Framework Extensibility}\label{sec:framework_extensibility}

We design NeMo-Aligner with extensibility in mind,  allowing users to easily modify algorithms in spite of the complexities of distributed training. We do so using the trainer abstraction, which encourages re-use of existing trainer methods across various steps and approaches. The extensibility of NeMo-Aligner allows variants of DPO to be integrated with minimal code changes, including the Identity Preference Optimization \citep{azar2023general}, the Conservative DPO \citep{mitchell2023note}, and the Kahneman-Tversky Optimization \citep{ethayarajh2023human}. Furthermore, other model alignment techniques such as Constitutional AI \citep{bai2022constitutional}, Rejection Sampling \citep{touvron2023llama}, and Self-Rewarding Language Models \citep{yuan2024selfrewarding} are also being incorporated into NeMo-Aligner, facilitated by the framework design.

\section{Conclusion}

Modern model alignment techniques, especially those based on Reinforcement Learning, pose complex optimization challenges with respect to system implementation.
We create and open-source NeMo-Aligner to allow AI researchers and practitioners to efficiently experiment with LLM alignment by utilizing all available compute in a scalable way. Our framework consistently scales well when training large models with more compute. 
As this is our initial release, we expect this scaling to only improve with future versions. Additionally, we support SFT, PPO, DPO, SteerLM in a parameter-efficient manner using LoRA for compute-limited settings. As an Apache 2.0 licensed open-source codebase, NeMo-Aligner can make alignment research more efficient and accessible.

\section*{Acknowledgements}
We would like to thank many teams at NVIDIA who contributed towards enabling NeMo-Aligner, especially the NeMo, TRT-LLM and TensorRT teams.
\bibliography{custom}

\begin{thebibliography}{48}
\providecommand{\natexlab}[1]{#1}
\providecommand{\url}[1]{\texttt{#1}}
\expandafter\ifx\csname urlstyle\endcsname\relax
  \providecommand{\doi}[1]{doi: #1}\else
  \providecommand{\doi}{doi: \begingroup \urlstyle{rm}\Url}\fi

\bibitem[Azar et~al.(2023)Azar, Rowland, Piot, Guo, Calandriello, Valko, and Munos]{azar2023general}
Mohammad~Gheshlaghi Azar, Mark Rowland, Bilal Piot, Daniel Guo, Daniele Calandriello, Michal Valko, and R{\'e}mi Munos.
\newblock A general theoretical paradigm to understand learning from human preferences.
\newblock \emph{arXiv preprint arXiv:2310.12036}, 2023.

\bibitem[Bai et~al.(2022{\natexlab{a}})Bai, Jones, Ndousse, Askell, Chen, DasSarma, Drain, Fort, Ganguli, Henighan, Joseph, Kadavath, Kernion, Conerly, El-Showk, Elhage, Hatfield-Dodds, Hernandez, Hume, Johnston, Kravec, Lovitt, Nanda, Olsson, Amodei, Brown, Clark, McCandlish, Olah, Mann, and Kaplan]{bai2022training}
Yuntao Bai, Andy Jones, Kamal Ndousse, Amanda Askell, Anna Chen, Nova DasSarma, Dawn Drain, Stanislav Fort, Deep Ganguli, Tom Henighan, Nicholas Joseph, Saurav Kadavath, Jackson Kernion, Tom Conerly, Sheer El-Showk, Nelson Elhage, Zac Hatfield-Dodds, Danny Hernandez, Tristan Hume, Scott Johnston, Shauna Kravec, Liane Lovitt, Neel Nanda, Catherine Olsson, Dario Amodei, Tom Brown, Jack Clark, Sam McCandlish, Chris Olah, Ben Mann, and Jared Kaplan.
\newblock Training a helpful and harmless assistant with reinforcement learning from human feedback, 2022{\natexlab{a}}.

\bibitem[Bai et~al.(2022{\natexlab{b}})Bai, Kadavath, Kundu, Askell, Kernion, Jones, Chen, Goldie, Mirhoseini, McKinnon, et~al.]{bai2022constitutional}
Yuntao Bai, Saurav Kadavath, Sandipan Kundu, Amanda Askell, Jackson Kernion, Andy Jones, Anna Chen, Anna Goldie, Azalia Mirhoseini, Cameron McKinnon, et~al.
\newblock Constitutional ai: Harmlessness from ai feedback.
\newblock \emph{arXiv preprint arXiv:2212.08073}, 2022{\natexlab{b}}.

\bibitem[Bradley \& Terry(1952)Bradley and Terry]{bradleyterry1952}
Ralph~A. Bradley and Milton~E. Terry.
\newblock The rank analysis of incomplete block designs --- {I. The} method of paired comparisons.
\newblock \emph{Biometrika}, 39:\penalty0 324--345, 1952.

\bibitem[Brown et~al.(2020)Brown, Mann, Ryder, Subbiah, Kaplan, Dhariwal, Neelakantan, Shyam, Sastry, Askell, Agarwal, Herbert-Voss, Krueger, Henighan, Child, Ramesh, Ziegler, Wu, Winter, Hesse, Chen, Sigler, Litwin, Gray, Chess, Clark, Berner, McCandlish, Radford, Sutskever, and Amodei]{brown2020language}
Tom~B. Brown, Benjamin Mann, Nick Ryder, Melanie Subbiah, Jared Kaplan, Prafulla Dhariwal, Arvind Neelakantan, Pranav Shyam, Girish Sastry, Amanda Askell, Sandhini Agarwal, Ariel Herbert-Voss, Gretchen Krueger, Tom Henighan, Rewon Child, Aditya Ramesh, Daniel~M. Ziegler, Jeffrey Wu, Clemens Winter, Christopher Hesse, Mark Chen, Eric Sigler, Mateusz Litwin, Scott Gray, Benjamin Chess, Jack Clark, Christopher Berner, Sam McCandlish, Alec Radford, Ilya Sutskever, and Dario Amodei.
\newblock Language models are few-shot learners, 2020.

\bibitem[Chen et~al.(2024)Chen, Deng, Yuan, Ji, and Gu]{chen2024selfplay}
Zixiang Chen, Yihe Deng, Huizhuo Yuan, Kaixuan Ji, and Quanquan Gu.
\newblock Self-play fine-tuning converts weak language models to strong language models, 2024.

\bibitem[Christiano et~al.(2017)Christiano, Leike, Brown, Martic, Legg, and Amodei]{christiano2017rlhf}
Paul~F Christiano, Jan Leike, Tom Brown, Miljan Martic, Shane Legg, and Dario Amodei.
\newblock Deep reinforcement learning from human preferences.
\newblock In I.~Guyon, U.~Von Luxburg, S.~Bengio, H.~Wallach, R.~Fergus, S.~Vishwanathan, and R.~Garnett (eds.), \emph{Advances in Neural Information Processing Systems}, volume~30. Curran Associates, Inc., 2017.

\bibitem[Conover et~al.(2023)Conover, Hayes, Mathur, Meng, Xie, Wan, Shah, Ghodsi, Wendell, Zaharia, and et~al.]{Conover_Hayes_2023}
Mike Conover, Matt Hayes, Ankit Mathur, Xiangrui Meng, Jianwei Xie, Jun Wan, Sam Shah, Ali Ghodsi, Patrick Wendell, Matei Zaharia, and et~al.
\newblock Free dolly: Introducing the world’s first truly open instruction-tuned llm, 2023.
\newblock URL \url{https://www.databricks.com/blog/2023/04/12/dolly-first-open-commercially-viable-instruction-tuned-llm}.

\bibitem[Cui et~al.(2023)Cui, Yuan, Ding, Yao, Zhu, Ni, Xie, Liu, and Sun]{cui2023ultrafeedback}
Ganqu Cui, Lifan Yuan, Ning Ding, Guanming Yao, Wei Zhu, Yuan Ni, Guotong Xie, Zhiyuan Liu, and Maosong Sun.
\newblock Ultrafeedback: Boosting language models with high-quality feedback.
\newblock \emph{arXiv preprint arXiv:2310.01377}, 2023.

\bibitem[Ding et~al.(2023)Ding, Chen, Xu, Qin, Zheng, Hu, Liu, Sun, and Zhou]{ding2023enhancing}
Ning Ding, Yulin Chen, Bokai Xu, Yujia Qin, Zhi Zheng, Shengding Hu, Zhiyuan Liu, Maosong Sun, and Bowen Zhou.
\newblock Enhancing chat language models by scaling high-quality instructional conversations.
\newblock \emph{arXiv preprint arXiv:2305.14233}, 2023.

\bibitem[Dong et~al.(2023)Dong, Wang, Sreedhar, Wu, and Kuchaiev]{dong-etal-2023-steerlm}
Yi~Dong, Zhilin Wang, Makesh Sreedhar, Xianchao Wu, and Oleksii Kuchaiev.
\newblock {S}teer{LM}: Attribute conditioned {SFT} as an (user-steerable) alternative to {RLHF}.
\newblock In Houda Bouamor, Juan Pino, and Kalika Bali (eds.), \emph{Findings of the Association for Computational Linguistics: EMNLP 2023}, pp.\  11275--11288, Singapore, December 2023. Association for Computational Linguistics.
\newblock \doi{10.18653/v1/2023.findings-emnlp.754}.
\newblock URL \url{https://aclanthology.org/2023.findings-emnlp.754}.

\bibitem[Dubey et~al.(2024)Dubey, Jauhri, Pandey, Kadian, Al-Dahle, Letman, Mathur, Schelten, Yang, Fan, Goyal, Hartshorn, Yang, Mitra, Sravankumar, Korenev, Hinsvark, Rao, Zhang, Rodriguez, Gregerson, Spataru, Roziere, Biron, Tang, Chern, Caucheteux, Nayak, Bi, Marra, McConnell, Keller, Touret, Wu, Wong, Ferrer, Nikolaidis, Allonsius, Song, Pintz, Livshits, Esiobu, Choudhary, Mahajan, Garcia-Olano, Perino, Hupkes, Lakomkin, AlBadawy, Lobanova, Dinan, Smith, Radenovic, Zhang, Synnaeve, Lee, Anderson, Nail, Mialon, Pang, Cucurell, Nguyen, Korevaar, Xu, Touvron, Zarov, Ibarra, Kloumann, Misra, Evtimov, Copet, Lee, Geffert, Vranes, Park, Mahadeokar, Shah, van~der Linde, Billock, Hong, Lee, Fu, Chi, Huang, Liu, Wang, Yu, Bitton, Spisak, Park, Rocca, Johnstun, Saxe, Jia, Alwala, Upasani, Plawiak, Li, Heafield, Stone, El-Arini, Iyer, Malik, Chiu, Bhalla, Rantala-Yeary, van~der Maaten, Chen, Tan, Jenkins, Martin, Madaan, Malo, Blecher, Landzaat, de~Oliveira, Muzzi, Pasupuleti, Singh, Paluri, Kardas, Oldham, Rita,
  Pavlova, Kambadur, Lewis, Si, Singh, Hassan, Goyal, Torabi, Bashlykov, Bogoychev, Chatterji, Duchenne, Çelebi, Alrassy, Zhang, Li, Vasic, Weng, Bhargava, Dubal, Krishnan, Koura, Xu, He, Dong, Srinivasan, Ganapathy, Calderer, Cabral, Stojnic, Raileanu, Girdhar, Patel, Sauvestre, Polidoro, Sumbaly, Taylor, Silva, Hou, Wang, Hosseini, Chennabasappa, Singh, Bell, Kim, Edunov, Nie, Narang, Raparthy, Shen, Wan, Bhosale, Zhang, Vandenhende, Batra, Whitman, Sootla, Collot, Gururangan, Borodinsky, Herman, Fowler, Sheasha, Georgiou, Scialom, Speckbacher, Mihaylov, Xiao, Karn, Goswami, Gupta, Ramanathan, Kerkez, Gonguet, Do, Vogeti, Petrovic, Chu, Xiong, Fu, Meers, Martinet, Wang, Tan, Xie, Jia, Wang, Goldschlag, Gaur, Babaei, Wen, Song, Zhang, Li, Mao, Coudert, Yan, Chen, Papakipos, Singh, Grattafiori, Jain, Kelsey, Shajnfeld, Gangidi, Victoria, Goldstand, Menon, Sharma, Boesenberg, Vaughan, Baevski, Feinstein, Kallet, Sangani, Yunus, Lupu, Alvarado, Caples, Gu, Ho, Poulton, Ryan, Ramchandani, Franco, Saraf,
  Chowdhury, Gabriel, Bharambe, Eisenman, Yazdan, James, Maurer, Leonhardi, Huang, Loyd, Paola, Paranjape, Liu, Wu, Ni, Hancock, Wasti, Spence, Stojkovic, Gamido, Montalvo, Parker, Burton, Mejia, Wang, Kim, Zhou, Hu, Chu, Cai, Tindal, Feichtenhofer, Civin, Beaty, Kreymer, Li, Wyatt, Adkins, Xu, Testuggine, David, Parikh, Liskovich, Foss, Wang, Le, Holland, Dowling, Jamil, Montgomery, Presani, Hahn, Wood, Brinkman, Arcaute, Dunbar, Smothers, Sun, Kreuk, Tian, Ozgenel, Caggioni, Guzmán, Kanayet, Seide, Florez, Schwarz, Badeer, Swee, Halpern, Thattai, Herman, Sizov, Guangyi, Zhang, Lakshminarayanan, Shojanazeri, Zou, Wang, Zha, Habeeb, Rudolph, Suk, Aspegren, Goldman, Molybog, Tufanov, Veliche, Gat, Weissman, Geboski, Kohli, Asher, Gaya, Marcus, Tang, Chan, Zhen, Reizenstein, Teboul, Zhong, Jin, Yang, Cummings, Carvill, Shepard, McPhie, Torres, Ginsburg, Wang, Wu, U, Saxena, Prasad, Khandelwal, Zand, Matosich, Veeraraghavan, Michelena, Li, Huang, Chawla, Lakhotia, Huang, Chen, Garg, A, Silva, Bell, Zhang, Guo,
  Yu, Moshkovich, Wehrstedt, Khabsa, Avalani, Bhatt, Tsimpoukelli, Mankus, Hasson, Lennie, Reso, Groshev, Naumov, Lathi, Keneally, Seltzer, Valko, Restrepo, Patel, Vyatskov, Samvelyan, Clark, Macey, Wang, Hermoso, Metanat, Rastegari, Bansal, Santhanam, Parks, White, Bawa, Singhal, Egebo, Usunier, Laptev, Dong, Zhang, Cheng, Chernoguz, Hart, Salpekar, Kalinli, Kent, Parekh, Saab, Balaji, Rittner, Bontrager, Roux, Dollar, Zvyagina, Ratanchandani, Yuvraj, Liang, Alao, Rodriguez, Ayub, Murthy, Nayani, Mitra, Li, Hogan, Battey, Wang, Maheswari, Howes, Rinott, Bondu, Datta, Chugh, Hunt, Dhillon, Sidorov, Pan, Verma, Yamamoto, Ramaswamy, Lindsay, Lindsay, Feng, Lin, Zha, Shankar, Zhang, Zhang, Wang, Agarwal, Sajuyigbe, Chintala, Max, Chen, Kehoe, Satterfield, Govindaprasad, Gupta, Cho, Virk, Subramanian, Choudhury, Goldman, Remez, Glaser, Best, Kohler, Robinson, Li, Zhang, Matthews, Chou, Shaked, Vontimitta, Ajayi, Montanez, Mohan, Kumar, Mangla, Ionescu, Poenaru, Mihailescu, Ivanov, Li, Wang, Jiang, Bouaziz,
  Constable, Tang, Wang, Wu, Wang, Xia, Wu, Gao, Chen, Hu, Jia, Qi, Li, Zhang, Zhang, Adi, Nam, Yu, Wang, Hao, Qian, He, Rait, DeVito, Rosnbrick, Wen, Yang, and Zhao]{llama3.1}
Abhimanyu Dubey, Abhinav Jauhri, Abhinav Pandey, Abhishek Kadian, Ahmad Al-Dahle, Aiesha Letman, Akhil Mathur, Alan Schelten, Amy Yang, Angela Fan, Anirudh Goyal, Anthony Hartshorn, Aobo Yang, Archi Mitra, Archie Sravankumar, Artem Korenev, Arthur Hinsvark, Arun Rao, Aston Zhang, Aurelien Rodriguez, Austen Gregerson, Ava Spataru, Baptiste Roziere, Bethany Biron, Binh Tang, Bobbie Chern, Charlotte Caucheteux, Chaya Nayak, Chloe Bi, Chris Marra, Chris McConnell, Christian Keller, Christophe Touret, Chunyang Wu, Corinne Wong, Cristian~Canton Ferrer, Cyrus Nikolaidis, Damien Allonsius, Daniel Song, Danielle Pintz, Danny Livshits, David Esiobu, Dhruv Choudhary, Dhruv Mahajan, Diego Garcia-Olano, Diego Perino, Dieuwke Hupkes, Egor Lakomkin, Ehab AlBadawy, Elina Lobanova, Emily Dinan, Eric~Michael Smith, Filip Radenovic, Frank Zhang, Gabriel Synnaeve, Gabrielle Lee, Georgia~Lewis Anderson, Graeme Nail, Gregoire Mialon, Guan Pang, Guillem Cucurell, Hailey Nguyen, Hannah Korevaar, Hu~Xu, Hugo Touvron, Iliyan Zarov,
  Imanol~Arrieta Ibarra, Isabel Kloumann, Ishan Misra, Ivan Evtimov, Jade Copet, Jaewon Lee, Jan Geffert, Jana Vranes, Jason Park, Jay Mahadeokar, Jeet Shah, Jelmer van~der Linde, Jennifer Billock, Jenny Hong, Jenya Lee, Jeremy Fu, Jianfeng Chi, Jianyu Huang, Jiawen Liu, Jie Wang, Jiecao Yu, Joanna Bitton, Joe Spisak, Jongsoo Park, Joseph Rocca, Joshua Johnstun, Joshua Saxe, Junteng Jia, Kalyan~Vasuden Alwala, Kartikeya Upasani, Kate Plawiak, Ke~Li, Kenneth Heafield, Kevin Stone, Khalid El-Arini, Krithika Iyer, Kshitiz Malik, Kuenley Chiu, Kunal Bhalla, Lauren Rantala-Yeary, Laurens van~der Maaten, Lawrence Chen, Liang Tan, Liz Jenkins, Louis Martin, Lovish Madaan, Lubo Malo, Lukas Blecher, Lukas Landzaat, Luke de~Oliveira, Madeline Muzzi, Mahesh Pasupuleti, Mannat Singh, Manohar Paluri, Marcin Kardas, Mathew Oldham, Mathieu Rita, Maya Pavlova, Melanie Kambadur, Mike Lewis, Min Si, Mitesh~Kumar Singh, Mona Hassan, Naman Goyal, Narjes Torabi, Nikolay Bashlykov, Nikolay Bogoychev, Niladri Chatterji, Olivier
  Duchenne, Onur Çelebi, Patrick Alrassy, Pengchuan Zhang, Pengwei Li, Petar Vasic, Peter Weng, Prajjwal Bhargava, Pratik Dubal, Praveen Krishnan, Punit~Singh Koura, Puxin Xu, Qing He, Qingxiao Dong, Ragavan Srinivasan, Raj Ganapathy, Ramon Calderer, Ricardo~Silveira Cabral, Robert Stojnic, Roberta Raileanu, Rohit Girdhar, Rohit Patel, Romain Sauvestre, Ronnie Polidoro, Roshan Sumbaly, Ross Taylor, Ruan Silva, Rui Hou, Rui Wang, Saghar Hosseini, Sahana Chennabasappa, Sanjay Singh, Sean Bell, Seohyun~Sonia Kim, Sergey Edunov, Shaoliang Nie, Sharan Narang, Sharath Raparthy, Sheng Shen, Shengye Wan, Shruti Bhosale, Shun Zhang, Simon Vandenhende, Soumya Batra, Spencer Whitman, Sten Sootla, Stephane Collot, Suchin Gururangan, Sydney Borodinsky, Tamar Herman, Tara Fowler, Tarek Sheasha, Thomas Georgiou, Thomas Scialom, Tobias Speckbacher, Todor Mihaylov, Tong Xiao, Ujjwal Karn, Vedanuj Goswami, Vibhor Gupta, Vignesh Ramanathan, Viktor Kerkez, Vincent Gonguet, Virginie Do, Vish Vogeti, Vladan Petrovic, Weiwei Chu,
  Wenhan Xiong, Wenyin Fu, Whitney Meers, Xavier Martinet, Xiaodong Wang, Xiaoqing~Ellen Tan, Xinfeng Xie, Xuchao Jia, Xuewei Wang, Yaelle Goldschlag, Yashesh Gaur, Yasmine Babaei, Yi~Wen, Yiwen Song, Yuchen Zhang, Yue Li, Yuning Mao, Zacharie~Delpierre Coudert, Zheng Yan, Zhengxing Chen, Zoe Papakipos, Aaditya Singh, Aaron Grattafiori, Abha Jain, Adam Kelsey, Adam Shajnfeld, Adithya Gangidi, Adolfo Victoria, Ahuva Goldstand, Ajay Menon, Ajay Sharma, Alex Boesenberg, Alex Vaughan, Alexei Baevski, Allie Feinstein, Amanda Kallet, Amit Sangani, Anam Yunus, Andrei Lupu, Andres Alvarado, Andrew Caples, Andrew Gu, Andrew Ho, Andrew Poulton, Andrew Ryan, Ankit Ramchandani, Annie Franco, Aparajita Saraf, Arkabandhu Chowdhury, Ashley Gabriel, Ashwin Bharambe, Assaf Eisenman, Azadeh Yazdan, Beau James, Ben Maurer, Benjamin Leonhardi, Bernie Huang, Beth Loyd, Beto~De Paola, Bhargavi Paranjape, Bing Liu, Bo~Wu, Boyu Ni, Braden Hancock, Bram Wasti, Brandon Spence, Brani Stojkovic, Brian Gamido, Britt Montalvo, Carl
  Parker, Carly Burton, Catalina Mejia, Changhan Wang, Changkyu Kim, Chao Zhou, Chester Hu, Ching-Hsiang Chu, Chris Cai, Chris Tindal, Christoph Feichtenhofer, Damon Civin, Dana Beaty, Daniel Kreymer, Daniel Li, Danny Wyatt, David Adkins, David Xu, Davide Testuggine, Delia David, Devi Parikh, Diana Liskovich, Didem Foss, Dingkang Wang, Duc Le, Dustin Holland, Edward Dowling, Eissa Jamil, Elaine Montgomery, Eleonora Presani, Emily Hahn, Emily Wood, Erik Brinkman, Esteban Arcaute, Evan Dunbar, Evan Smothers, Fei Sun, Felix Kreuk, Feng Tian, Firat Ozgenel, Francesco Caggioni, Francisco Guzmán, Frank Kanayet, Frank Seide, Gabriela~Medina Florez, Gabriella Schwarz, Gada Badeer, Georgia Swee, Gil Halpern, Govind Thattai, Grant Herman, Grigory Sizov, Guangyi, Zhang, Guna Lakshminarayanan, Hamid Shojanazeri, Han Zou, Hannah Wang, Hanwen Zha, Haroun Habeeb, Harrison Rudolph, Helen Suk, Henry Aspegren, Hunter Goldman, Igor Molybog, Igor Tufanov, Irina-Elena Veliche, Itai Gat, Jake Weissman, James Geboski, James Kohli,
  Japhet Asher, Jean-Baptiste Gaya, Jeff Marcus, Jeff Tang, Jennifer Chan, Jenny Zhen, Jeremy Reizenstein, Jeremy Teboul, Jessica Zhong, Jian Jin, Jingyi Yang, Joe Cummings, Jon Carvill, Jon Shepard, Jonathan McPhie, Jonathan Torres, Josh Ginsburg, Junjie Wang, Kai Wu, Kam~Hou U, Karan Saxena, Karthik Prasad, Kartikay Khandelwal, Katayoun Zand, Kathy Matosich, Kaushik Veeraraghavan, Kelly Michelena, Keqian Li, Kun Huang, Kunal Chawla, Kushal Lakhotia, Kyle Huang, Lailin Chen, Lakshya Garg, Lavender A, Leandro Silva, Lee Bell, Lei Zhang, Liangpeng Guo, Licheng Yu, Liron Moshkovich, Luca Wehrstedt, Madian Khabsa, Manav Avalani, Manish Bhatt, Maria Tsimpoukelli, Martynas Mankus, Matan Hasson, Matthew Lennie, Matthias Reso, Maxim Groshev, Maxim Naumov, Maya Lathi, Meghan Keneally, Michael~L. Seltzer, Michal Valko, Michelle Restrepo, Mihir Patel, Mik Vyatskov, Mikayel Samvelyan, Mike Clark, Mike Macey, Mike Wang, Miquel~Jubert Hermoso, Mo~Metanat, Mohammad Rastegari, Munish Bansal, Nandhini Santhanam, Natascha
  Parks, Natasha White, Navyata Bawa, Nayan Singhal, Nick Egebo, Nicolas Usunier, Nikolay~Pavlovich Laptev, Ning Dong, Ning Zhang, Norman Cheng, Oleg Chernoguz, Olivia Hart, Omkar Salpekar, Ozlem Kalinli, Parkin Kent, Parth Parekh, Paul Saab, Pavan Balaji, Pedro Rittner, Philip Bontrager, Pierre Roux, Piotr Dollar, Polina Zvyagina, Prashant Ratanchandani, Pritish Yuvraj, Qian Liang, Rachad Alao, Rachel Rodriguez, Rafi Ayub, Raghotham Murthy, Raghu Nayani, Rahul Mitra, Raymond Li, Rebekkah Hogan, Robin Battey, Rocky Wang, Rohan Maheswari, Russ Howes, Ruty Rinott, Sai~Jayesh Bondu, Samyak Datta, Sara Chugh, Sara Hunt, Sargun Dhillon, Sasha Sidorov, Satadru Pan, Saurabh Verma, Seiji Yamamoto, Sharadh Ramaswamy, Shaun Lindsay, Shaun Lindsay, Sheng Feng, Shenghao Lin, Shengxin~Cindy Zha, Shiva Shankar, Shuqiang Zhang, Shuqiang Zhang, Sinong Wang, Sneha Agarwal, Soji Sajuyigbe, Soumith Chintala, Stephanie Max, Stephen Chen, Steve Kehoe, Steve Satterfield, Sudarshan Govindaprasad, Sumit Gupta, Sungmin Cho, Sunny
  Virk, Suraj Subramanian, Sy~Choudhury, Sydney Goldman, Tal Remez, Tamar Glaser, Tamara Best, Thilo Kohler, Thomas Robinson, Tianhe Li, Tianjun Zhang, Tim Matthews, Timothy Chou, Tzook Shaked, Varun Vontimitta, Victoria Ajayi, Victoria Montanez, Vijai Mohan, Vinay~Satish Kumar, Vishal Mangla, Vlad Ionescu, Vlad Poenaru, Vlad~Tiberiu Mihailescu, Vladimir Ivanov, Wei Li, Wenchen Wang, Wenwen Jiang, Wes Bouaziz, Will Constable, Xiaocheng Tang, Xiaofang Wang, Xiaojian Wu, Xiaolan Wang, Xide Xia, Xilun Wu, Xinbo Gao, Yanjun Chen, Ye~Hu, Ye~Jia, Ye~Qi, Yenda Li, Yilin Zhang, Ying Zhang, Yossi Adi, Youngjin Nam, Yu, Wang, Yuchen Hao, Yundi Qian, Yuzi He, Zach Rait, Zachary DeVito, Zef Rosnbrick, Zhaoduo Wen, Zhenyu Yang, and Zhiwei Zhao.
\newblock The llama 3 herd of models, 2024.
\newblock URL \url{https://arxiv.org/abs/2407.21783}.

\bibitem[Ethayarajh et~al.(2023)Ethayarajh, Xu, Jurafsky, and Kiela]{ethayarajh2023human}
Kawin Ethayarajh, Winnie Xu, Dan Jurafsky, and Douwe Kiela.
\newblock human-centerred loss functions (halos).
\newblock \url{https://github.com/ContextualAI/HALOs/blob/main/assets/report.pdf}, 2023.

\bibitem[Havrilla et~al.(2023)Havrilla, Zhuravinskyi, Phung, Tiwari, Tow, Biderman, Anthony, and Castricato]{havrilla-etal-2023-trlx}
Alexander Havrilla, Maksym Zhuravinskyi, Duy Phung, Aman Tiwari, Jonathan Tow, Stella Biderman, Quentin Anthony, and Louis Castricato.
\newblock trl{X}: A framework for large scale reinforcement learning from human feedback.
\newblock In \emph{Proceedings of the 2023 Conference on Empirical Methods in Natural Language Processing}, pp.\  8578--8595, Singapore, December 2023. Association for Computational Linguistics.
\newblock \doi{10.18653/v1/2023.emnlp-main.530}.
\newblock URL \url{https://aclanthology.org/2023.emnlp-main.530}.

\bibitem[Hu et~al.(2021)Hu, Shen, Wallis, Allen-Zhu, Li, Wang, Wang, and Chen]{hu2021lora}
Edward~J Hu, Yelong Shen, Phillip Wallis, Zeyuan Allen-Zhu, Yuanzhi Li, Shean Wang, Lu~Wang, and Weizhu Chen.
\newblock Lora: Low-rank adaptation of large language models.
\newblock \emph{arXiv preprint arXiv:2106.09685}, 2021.

\bibitem[Jiang et~al.(2023)Jiang, Sablayrolles, Mensch, Bamford, Chaplot, de~las Casas, Bressand, Lengyel, Lample, Saulnier, Lavaud, Lachaux, Stock, Scao, Lavril, Wang, Lacroix, and Sayed]{jiang2023mistral}
Albert~Q. Jiang, Alexandre Sablayrolles, Arthur Mensch, Chris Bamford, Devendra~Singh Chaplot, Diego de~las Casas, Florian Bressand, Gianna Lengyel, Guillaume Lample, Lucile Saulnier, Lélio~Renard Lavaud, Marie-Anne Lachaux, Pierre Stock, Teven~Le Scao, Thibaut Lavril, Thomas Wang, Timothée Lacroix, and William~El Sayed.
\newblock Mistral 7b, 2023.

\bibitem[Kuchaiev et~al.(2019)Kuchaiev, Li, Nguyen, Hrinchuk, Leary, Ginsburg, Kriman, Beliaev, Lavrukhin, Cook, Castonguay, Popova, Huang, and Cohen]{nemo}
Oleksii Kuchaiev, Jason Li, Huyen Nguyen, Oleksii Hrinchuk, Ryan Leary, Boris Ginsburg, Samuel Kriman, Stanislav Beliaev, Vitaly Lavrukhin, Jack Cook, Patrice Castonguay, Mariya Popova, Jocelyn Huang, and Jonathan~M. Cohen.
\newblock Nemo: a toolkit for building ai applications using neural modules, 2019.

\bibitem[Köpf et~al.(2023)Köpf, Kilcher, von Rütte, Anagnostidis, Tam, Stevens, Barhoum, Duc, Stanley, Nagyfi, ES, Suri, Glushkov, Dantuluri, Maguire, Schuhmann, Nguyen, and Mattick]{kopf2023openassistant}
Andreas Köpf, Yannic Kilcher, Dimitri von Rütte, Sotiris Anagnostidis, Zhi-Rui Tam, Keith Stevens, Abdullah Barhoum, Nguyen~Minh Duc, Oliver Stanley, Richárd Nagyfi, Shahul ES, Sameer Suri, David Glushkov, Arnav Dantuluri, Andrew Maguire, Christoph Schuhmann, Huu Nguyen, and Alexander Mattick.
\newblock Openassistant conversations -- democratizing large language model alignment, 2023.

\bibitem[Lambert \& Calandra(2023)Lambert and Calandra]{lambert2023alignment}
Nathan Lambert and Roberto Calandra.
\newblock The alignment ceiling: Objective mismatch in reinforcement learning from human feedback, 2023.

\bibitem[Meng et~al.(2024)Meng, Xia, and Chen]{meng2024simposimplepreferenceoptimization}
Yu~Meng, Mengzhou Xia, and Danqi Chen.
\newblock Simpo: Simple preference optimization with a reference-free reward, 2024.
\newblock URL \url{https://arxiv.org/abs/2405.14734}.

\bibitem[{Meta AI}(2024)]{llama3modelcard}
{Meta AI}.
\newblock Llama 3 model card.
\newblock \url{https://github.com/meta-llama/llama3/blob/main/MODEL_CARD.md}, 2024.

\bibitem[Mitchell(2023)]{mitchell2023note}
Eric Mitchell.
\newblock A note on {DPO} with noisy preferences \& relationship to {IPO}.
\newblock \url{https://ericmitchell.ai/cdpo.pdf}, 2023.

\bibitem[{NVIDIA}(2022)]{pytriton}
{NVIDIA}.
\newblock {PyTriton}: Framework facilitating {NVIDIA Triton} inference server usage in {Python} environments, 2022.
\newblock URL \url{https://github.com/triton-inference-server/pytriton}.

\bibitem[NVIDIA(2022)]{te}
NVIDIA.
\newblock {TransformerEngine}, 2022.
\newblock URL \url{https://github.com/NVIDIA/TransformerEngine}.

\bibitem[NVIDIA(2023{\natexlab{a}})]{nvlink}
NVIDIA.
\newblock {NVLink}.
\newblock \url{https://blogs.nvidia.com/blog/what-is-nvidia-nvlink/}, 2023{\natexlab{a}}.

\bibitem[NVIDIA(2023{\natexlab{b}})]{trtllm}
NVIDIA.
\newblock {TensorRT-LLM}.
\newblock \url{https://github.com/NVIDIA/TensorRT-LLM}, 2023{\natexlab{b}}.

\bibitem[NVIDIA(2023{\natexlab{c}})]{trtrefitter}
NVIDIA.
\newblock {TensorRT Refitter}.
\newblock \url{https://docs.nvidia.com/deeplearning/tensorrt/api/python_api/infer/Core/Refitter.html}, 2023{\natexlab{c}}.

\bibitem[NVIDIA(2024)]{infiniband}
NVIDIA.
\newblock {Infiniband}.
\newblock \url{https://www.nvidia.com/en-us/networking/products/infiniband/}, 2024.

\bibitem[Nvidia et~al.(2024)Nvidia, :, Adler, Agarwal, Aithal, Anh, Bhattacharya, Brundyn, Casper, Catanzaro, Clay, Cohen, Das, Dattagupta, Delalleau, Derczynski, Dong, Egert, Evans, Ficek, Fridman, Ghosh, Ginsburg, Gitman, Grzegorzek, Hero, Huang, Jawa, Jennings, Jhunjhunwala, Kamalu, Khan, Kuchaiev, LeGresley, Li, Liu, Liu, Long, Mahabaleshwarkar, Majumdar, Maki, Martinez, de~Melo, Moshkov, Narayanan, Narenthiran, Navarro, Nguyen, Nitski, Noroozi, Nutheti, Parisien, Parmar, Patwary, Pawelec, Ping, Prabhumoye, Roy, Saar, Sabavat, Satheesh, Scowcroft, Sewall, Shamis, Shen, Shoeybi, Sizer, Smelyanskiy, Soares, Sreedhar, Su, Subramanian, Sun, Toshniwal, Wang, Wang, You, Zeng, Zhang, Zhang, Zhang, Zhang, and Zhu]{nvidia2024nemotron4340btechnicalreport}
Nvidia, :, Bo~Adler, Niket Agarwal, Ashwath Aithal, Dong~H. Anh, Pallab Bhattacharya, Annika Brundyn, Jared Casper, Bryan Catanzaro, Sharon Clay, Jonathan Cohen, Sirshak Das, Ayush Dattagupta, Olivier Delalleau, Leon Derczynski, Yi~Dong, Daniel Egert, Ellie Evans, Aleksander Ficek, Denys Fridman, Shaona Ghosh, Boris Ginsburg, Igor Gitman, Tomasz Grzegorzek, Robert Hero, Jining Huang, Vibhu Jawa, Joseph Jennings, Aastha Jhunjhunwala, John Kamalu, Sadaf Khan, Oleksii Kuchaiev, Patrick LeGresley, Hui Li, Jiwei Liu, Zihan Liu, Eileen Long, Ameya~Sunil Mahabaleshwarkar, Somshubra Majumdar, James Maki, Miguel Martinez, Maer~Rodrigues de~Melo, Ivan Moshkov, Deepak Narayanan, Sean Narenthiran, Jesus Navarro, Phong Nguyen, Osvald Nitski, Vahid Noroozi, Guruprasad Nutheti, Christopher Parisien, Jupinder Parmar, Mostofa Patwary, Krzysztof Pawelec, Wei Ping, Shrimai Prabhumoye, Rajarshi Roy, Trisha Saar, Vasanth Rao~Naik Sabavat, Sanjeev Satheesh, Jane~Polak Scowcroft, Jason Sewall, Pavel Shamis, Gerald Shen, Mohammad
  Shoeybi, Dave Sizer, Misha Smelyanskiy, Felipe Soares, Makesh~Narsimhan Sreedhar, Dan Su, Sandeep Subramanian, Shengyang Sun, Shubham Toshniwal, Hao Wang, Zhilin Wang, Jiaxuan You, Jiaqi Zeng, Jimmy Zhang, Jing Zhang, Vivienne Zhang, Yian Zhang, and Chen Zhu.
\newblock Nemotron-4 340b technical report, 2024.
\newblock URL \url{https://arxiv.org/abs/2406.11704}.

\bibitem[Ouyang et~al.(2022)Ouyang, Wu, Jiang, Almeida, Wainwright, Mishkin, Zhang, Agarwal, Slama, Ray, Schulman, Hilton, Kelton, Miller, Simens, Askell, Welinder, Christiano, Leike, and Lowe]{instructgpt}
Long Ouyang, Jeffrey Wu, Xu~Jiang, Diogo Almeida, Carroll Wainwright, Pamela Mishkin, Chong Zhang, Sandhini Agarwal, Katarina Slama, Alex Ray, John Schulman, Jacob Hilton, Fraser Kelton, Luke Miller, Maddie Simens, Amanda Askell, Peter Welinder, Paul~F Christiano, Jan Leike, and Ryan Lowe.
\newblock Training language models to follow instructions with human feedback.
\newblock In S.~Koyejo, S.~Mohamed, A.~Agarwal, D.~Belgrave, K.~Cho, and A.~Oh (eds.), \emph{Advances in Neural Information Processing Systems}, volume~35, pp.\  27730--27744. Curran Associates, Inc., 2022.
\newblock URL \url{https://proceedings.neurips.cc/paper_files/paper/2022/file/b1efde53be364a73914f58805a001731-Paper-Conference.pdf}.

\bibitem[Rafailov et~al.(2023)Rafailov, Sharma, Mitchell, Ermon, Manning, and Finn]{rafailov2023direct}
Rafael Rafailov, Archit Sharma, Eric Mitchell, Stefano Ermon, Christopher~D. Manning, and Chelsea Finn.
\newblock Direct preference optimization: Your language model is secretly a reward model, 2023.

\bibitem[Sanh et~al.(2022)Sanh, Webson, Raffel, Bach, Sutawika, Alyafeai, Chaffin, Stiegler, Scao, Raja, Dey, Bari, Xu, Thakker, Sharma, Szczechla, Kim, Chhablani, Nayak, Datta, Chang, Jiang, Wang, Manica, Shen, Yong, Pandey, Bawden, Wang, Neeraj, Rozen, Sharma, Santilli, Fevry, Fries, Teehan, Bers, Biderman, Gao, Wolf, and Rush]{sanh2022multitask}
Victor Sanh, Albert Webson, Colin Raffel, Stephen~H. Bach, Lintang Sutawika, Zaid Alyafeai, Antoine Chaffin, Arnaud Stiegler, Teven~Le Scao, Arun Raja, Manan Dey, M~Saiful Bari, Canwen Xu, Urmish Thakker, Shanya~Sharma Sharma, Eliza Szczechla, Taewoon Kim, Gunjan Chhablani, Nihal Nayak, Debajyoti Datta, Jonathan Chang, Mike Tian-Jian Jiang, Han Wang, Matteo Manica, Sheng Shen, Zheng~Xin Yong, Harshit Pandey, Rachel Bawden, Thomas Wang, Trishala Neeraj, Jos Rozen, Abheesht Sharma, Andrea Santilli, Thibault Fevry, Jason~Alan Fries, Ryan Teehan, Tali Bers, Stella Biderman, Leo Gao, Thomas Wolf, and Alexander~M. Rush.
\newblock Multitask prompted training enables zero-shot task generalization, 2022.

\bibitem[Schulman et~al.(2017)Schulman, Wolski, Dhariwal, Radford, and Klimov]{schulman2017proximal}
John Schulman, Filip Wolski, Prafulla Dhariwal, Alec Radford, and Oleg Klimov.
\newblock Proximal policy optimization algorithms, 2017.

\bibitem[Shoeybi et~al.(2020)Shoeybi, Patwary, Puri, LeGresley, Casper, and Catanzaro]{shoeybi2020megatronlm}
Mohammad Shoeybi, Mostofa Patwary, Raul Puri, Patrick LeGresley, Jared Casper, and Bryan Catanzaro.
\newblock Megatron-lm: Training multi-billion parameter language models using model parallelism, 2020.

\bibitem[Stiennon et~al.(2020)Stiennon, Ouyang, Wu, Ziegler, Lowe, Voss, Radford, Amodei, and Christiano]{stiennon2020summarize}
Nisan Stiennon, Long Ouyang, Jeff Wu, Daniel~M. Ziegler, Ryan Lowe, Chelsea Voss, Alec Radford, Dario Amodei, and Paul Christiano.
\newblock Learning to summarize from human feedback.
\newblock In \emph{Proceedings of the 34th International Conference on Neural Information Processing Systems}, NIPS'20, Red Hook, NY, USA, 2020. Curran Associates Inc.
\newblock ISBN 9781713829546.

\bibitem[Taori et~al.(2023)Taori, Gulrajani, Zhang, Dubois, Li, Guestrin, Liang, and Hashimoto]{alpaca}
Rohan Taori, Ishaan Gulrajani, Tianyi Zhang, Yann Dubois, Xuechen Li, Carlos Guestrin, Percy Liang, and Tatsunori~B. Hashimoto.
\newblock Stanford alpaca: An instruction-following llama model.
\newblock \url{https://github.com/tatsu-lab/stanford_alpaca}, 2023.

\bibitem[Touvron et~al.(2023)Touvron, Martin, Stone, Albert, Almahairi, Babaei, Bashlykov, Batra, Bhargava, Bhosale, Bikel, Blecher, Ferrer, Chen, Cucurull, Esiobu, Fernandes, Fu, Fu, Fuller, Gao, Goswami, Goyal, Hartshorn, Hosseini, Hou, Inan, Kardas, Kerkez, Khabsa, Kloumann, Korenev, Koura, Lachaux, Lavril, Lee, Liskovich, Lu, Mao, Martinet, Mihaylov, Mishra, Molybog, Nie, Poulton, Reizenstein, Rungta, Saladi, Schelten, Silva, Smith, Subramanian, Tan, Tang, Taylor, Williams, Kuan, Xu, Yan, Zarov, Zhang, Fan, Kambadur, Narang, Rodriguez, Stojnic, Edunov, and Scialom]{touvron2023llama}
Hugo Touvron, Louis Martin, Kevin Stone, Peter Albert, Amjad Almahairi, Yasmine Babaei, Nikolay Bashlykov, Soumya Batra, Prajjwal Bhargava, Shruti Bhosale, Dan Bikel, Lukas Blecher, Cristian~Canton Ferrer, Moya Chen, Guillem Cucurull, David Esiobu, Jude Fernandes, Jeremy Fu, Wenyin Fu, Brian Fuller, Cynthia Gao, Vedanuj Goswami, Naman Goyal, Anthony Hartshorn, Saghar Hosseini, Rui Hou, Hakan Inan, Marcin Kardas, Viktor Kerkez, Madian Khabsa, Isabel Kloumann, Artem Korenev, Punit~Singh Koura, Marie-Anne Lachaux, Thibaut Lavril, Jenya Lee, Diana Liskovich, Yinghai Lu, Yuning Mao, Xavier Martinet, Todor Mihaylov, Pushkar Mishra, Igor Molybog, Yixin Nie, Andrew Poulton, Jeremy Reizenstein, Rashi Rungta, Kalyan Saladi, Alan Schelten, Ruan Silva, Eric~Michael Smith, Ranjan Subramanian, Xiaoqing~Ellen Tan, Binh Tang, Ross Taylor, Adina Williams, Jian~Xiang Kuan, Puxin Xu, Zheng Yan, Iliyan Zarov, Yuchen Zhang, Angela Fan, Melanie Kambadur, Sharan Narang, Aurelien Rodriguez, Robert Stojnic, Sergey Edunov, and Thomas
  Scialom.
\newblock Llama 2: Open foundation and fine-tuned chat models, 2023.

\bibitem[Tunstall et~al.(2023)Tunstall, Beeching, Lambert, Rajani, Rasul, Belkada, Huang, von Werra, Fourrier, Habib, Sarrazin, Sanseviero, Rush, and Wolf]{tunstall2023zephyr}
Lewis Tunstall, Edward Beeching, Nathan Lambert, Nazneen Rajani, Kashif Rasul, Younes Belkada, Shengyi Huang, Leandro von Werra, Clémentine Fourrier, Nathan Habib, Nathan Sarrazin, Omar Sanseviero, Alexander~M. Rush, and Thomas Wolf.
\newblock Zephyr: Direct distillation of lm alignment, 2023.

\bibitem[von Werra et~al.(2020)von Werra, Belkada, Tunstall, Beeching, Thrush, Lambert, and Huang]{vonwerra2022trl}
Leandro von Werra, Younes Belkada, Lewis Tunstall, Edward Beeching, Tristan Thrush, Nathan Lambert, and Shengyi Huang.
\newblock Trl: Transformer reinforcement learning.
\newblock \url{https://github.com/huggingface/trl}, 2020.

\bibitem[Wang et~al.(2023)Wang, Dong, Zeng, Adams, Sreedhar, Egert, Delalleau, Scowcroft, Kant, Swope, and Kuchaiev]{wang2023helpsteer}
Zhilin Wang, Yi~Dong, Jiaqi Zeng, Virginia Adams, Makesh~Narsimhan Sreedhar, Daniel Egert, Olivier Delalleau, Jane~Polak Scowcroft, Neel Kant, Aidan Swope, and Oleksii Kuchaiev.
\newblock Helpsteer: Multi-attribute helpfulness dataset for steerlm, 2023.

\bibitem[Wang et~al.(2024)Wang, Dong, Delalleau, Zeng, Shen, Egert, Zhang, Sreedhar, and Kuchaiev]{helpsteer2}
Zhilin Wang, Yi~Dong, Olivier Delalleau, Jiaqi Zeng, Gerald Shen, Daniel Egert, Jimmy~J. Zhang, Makesh~Narsimhan Sreedhar, and Oleksii Kuchaiev.
\newblock Helpsteer2: Open-source dataset for training top-performing reward models, 2024.
\newblock URL \url{https://arxiv.org/abs/2406.08673}.

\bibitem[Wei et~al.(2022)Wei, Bosma, Zhao, Guu, Yu, Lester, Du, Dai, and Le]{wei2022finetuned}
Jason Wei, Maarten Bosma, Vincent~Y. Zhao, Kelvin Guu, Adams~Wei Yu, Brian Lester, Nan Du, Andrew~M. Dai, and Quoc~V. Le.
\newblock Finetuned language models are zero-shot learners, 2022.

\bibitem[Williams(1992)]{williams1992simple}
Ronald~J Williams.
\newblock Simple statistical gradient-following algorithms for connectionist reinforcement learning.
\newblock \emph{Machine learning}, 8:\penalty0 229--256, 1992.

\bibitem[Yao et~al.(2023)Yao, Aminabadi, Ruwase, Rajbhandari, Wu, Awan, Rasley, Zhang, Li, Holmes, Zhou, Wyatt, Smith, Kurilenko, Qin, Tanaka, Che, Song, and He]{yao2023dschat}
Zhewei Yao, Reza~Yazdani Aminabadi, Olatunji Ruwase, Samyam Rajbhandari, Xiaoxia Wu, Ammar~Ahmad Awan, Jeff Rasley, Minjia Zhang, Conglong Li, Connor Holmes, Zhongzhu Zhou, Michael Wyatt, Molly Smith, Lev Kurilenko, Heyang Qin, Masahiro Tanaka, Shuai Che, Shuaiwen~Leon Song, and Yuxiong He.
\newblock {DeepSpeed-Chat: Easy, Fast and Affordable RLHF Training of ChatGPT-like Models at All Scales}.
\newblock \emph{arXiv preprint arXiv:2308.01320}, 2023.

\bibitem[Yuan et~al.(2024)Yuan, Pang, Cho, Li, Sukhbaatar, Xu, and Weston]{yuan2024selfrewarding}
Weizhe Yuan, Richard~Yuanzhe Pang, Kyunghyun Cho, Xian Li, Sainbayar Sukhbaatar, Jing Xu, and Jason Weston.
\newblock Self-rewarding language models, 2024.

\bibitem[Zhang et~al.(2022)Zhang, Roller, Goyal, Artetxe, Chen, Chen, Dewan, Diab, Li, Lin, Mihaylov, Ott, Shleifer, Shuster, Simig, Koura, Sridhar, Wang, and Zettlemoyer]{zhang2022opt}
Susan Zhang, Stephen Roller, Naman Goyal, Mikel Artetxe, Moya Chen, Shuohui Chen, Christopher Dewan, Mona Diab, Xian Li, Xi~Victoria Lin, Todor Mihaylov, Myle Ott, Sam Shleifer, Kurt Shuster, Daniel Simig, Punit~Singh Koura, Anjali Sridhar, Tianlu Wang, and Luke Zettlemoyer.
\newblock Opt: Open pre-trained transformer language models, 2022.

\bibitem[Zheng et~al.(2023{\natexlab{a}})Zheng, Chiang, Sheng, Zhuang, Wu, Zhuang, Lin, Li, Li, Xing, et~al.]{zheng2023judging}
Lianmin Zheng, Wei-Lin Chiang, Ying Sheng, Siyuan Zhuang, Zhanghao Wu, Yonghao Zhuang, Zi~Lin, Zhuohan Li, Dacheng Li, Eric Xing, et~al.
\newblock Judging llm-as-a-judge with mt-bench and chatbot arena.
\newblock \emph{arXiv preprint arXiv:2306.05685}, 2023{\natexlab{a}}.

\bibitem[Zheng et~al.(2023{\natexlab{b}})Zheng, Dou, Gao, Hua, Shen, Wang, Liu, Jin, Liu, Zhou, Xiong, Chen, Xi, Xu, Lai, Zhu, Chang, Yin, Weng, Cheng, Huang, Sun, Yan, Gui, Zhang, Qiu, and Huang]{zheng2023secrets}
Rui Zheng, Shihan Dou, Songyang Gao, Yuan Hua, Wei Shen, Binghai Wang, Yan Liu, Senjie Jin, Qin Liu, Yuhao Zhou, Limao Xiong, Lu~Chen, Zhiheng Xi, Nuo Xu, Wenbin Lai, Minghao Zhu, Cheng Chang, Zhangyue Yin, Rongxiang Weng, Wensen Cheng, Haoran Huang, Tianxiang Sun, Hang Yan, Tao Gui, Qi~Zhang, Xipeng Qiu, and Xuanjing Huang.
\newblock Secrets of rlhf in large language models part i: Ppo, 2023{\natexlab{b}}.

\end{thebibliography}

\bibliographystyle{colm2024_conference}


\end{document}